\documentclass[10pt, conference]{IEEEtran}
\usepackage[utf8]{inputenc} 
\usepackage{graphicx,url}
\usepackage{multirow}
\usepackage{hhline}
\graphicspath{ {./images/} }
\usepackage{graphicx}
\usepackage{subfigure,amssymb,amsmath}
\usepackage{color,soul}
\usepackage[dvipsnames]{xcolor}
\usepackage{bbm}
\usepackage{tabularx}
\usepackage{float}
\usepackage{booktabs}       

\makeatletter
\newif\if@restonecol
\makeatother

\usepackage[linesnumbered,ruled,vlined]{algorithm2e}
\usepackage{algpseudocode}
\usepackage{amsmath}
\usepackage{fancyhdr} 
\usepackage{tabularx}
\usepackage{comment} 

\renewcommand\thetable{\arabic{table}}
\usepackage{tikz,pgfplots}
\usepackage{pgfplots}
\usetikzlibrary{shapes.geometric,backgrounds,patterns, trees}
\usetikzlibrary{3d,decorations.text,shapes.arrows,positioning,fit,backgrounds}
\usetikzlibrary{positioning, decorations.pathmorphing, shapes}
\usetikzlibrary{decorations.pathreplacing}
\usetikzlibrary{shapes.geometric,backgrounds,patterns, trees}
\usetikzlibrary{arrows.meta,
                bending,
                intersections,
                quotes,
                shapes.geometric}
              \usetikzlibrary{automata, positioning}
\usepgfplotslibrary{fillbetween}
\usetikzlibrary{shapes,arrows}
\usetikzlibrary{arrows.meta}
\usetikzlibrary{positioning}
\tikzset{set/.style={draw,circle,inner sep=0pt,align=center}}
\usetikzlibrary{automata, positioning}
  \usetikzlibrary{shapes,shadows}
  \tikzstyle{abstractbox} = [draw=black, fill=white, rectangle,
  inner sep=10pt, style=rounded corners, drop shadow={fill=black,
  opacity=1}]
\tikzstyle{abstracttitle} =[fill=white]
\tikzstyle{block} = [rectangle, draw,
text width=10.5em, text centered, rounded corners, minimum height=4em]
\tikzstyle{line} = [draw, -latex]
\usetikzlibrary{calc,positioning,shapes.geometric}
\usetikzlibrary{arrows.meta,arrows}

\usetikzlibrary{matrix}
\tikzstyle{cblue}=[circle, draw, thin,fill=cyan!20, scale=0.8]
\tikzstyle{qgre}=[rectangle, draw, thin,fill=green!20, scale=0.8]
\tikzstyle{rpath}=[ultra thick, red, opacity=0.4]
\tikzstyle{legend_isps}=[rectangle, rounded corners, thin,
                       fill=gray!20, text=blue, draw]

\tikzstyle{legend_overlay}=[rectangle, rounded corners, thin,
                           top color= white,bottom color=green!25,
                           minimum width=2.5cm, minimum height=0.8cm,
                           pinegreen]
\tikzstyle{legend_phytop}=[rectangle, rounded corners, thin,
                          top color= white,bottom color=cyan!25,
                          minimum width=2.5cm, minimum height=0.8cm,
                          royalblue]
\tikzstyle{legend_general}=[rectangle, rounded ckith'sers, thin,
                          top color= white,bottom color=lavander!25,
                          minimum width=2.5cm, minimum height=0.8cm,
                          violet]
                          \usetikzlibrary{matrix}

\colorlet{myRed}{red!20}
\tikzset{
  rows/.style 2 args={/utils/temp/.style={row ##1/.append style={nodes={#2}}},
    /utils/temp/.list={#1}},
  columns/.style 2 args={/utils/temp/.style={column ##1/.append style={nodes={#2}}},
    /utils/temp/.list={#1}}}
\usetikzlibrary{backgrounds,calc,shadings,shapes.arrows,shapes.symbols,shadows}
\definecolor{switch}{HTML}{006996}

\makeatletter
\pgfkeys{/pgf/.cd,
  parallelepiped offset x/.initial=2mm,
  parallelepiped offset y/.initial=2mm
}
\pgfdeclareshape{parallelepiped}
{
  \inheritsavedanchors[from=rectangle] 
  \inheritanchorborder[from=rectangle]
  \inheritanchor[from=rectangle]{north}
  \inheritanchor[from=rectangle]{north west}
  \inheritanchor[from=rectangle]{north east}
  \inheritanchor[from=rectangle]{center}
  \inheritanchor[from=rectangle]{west}
  \inheritanchor[from=rectangle]{east}
  \inheritanchor[from=rectangle]{mid}
  \inheritanchor[from=rectangle]{mid west}
  \inheritanchor[from=rectangle]{mid east}
  \inheritanchor[from=rectangle]{base}
  \inheritanchor[from=rectangle]{base west}
  \inheritanchor[from=rectangle]{base east}
  \inheritanchor[from=rectangle]{south}
  \inheritanchor[from=rectangle]{south west}
  \inheritanchor[from=rectangle]{south east}
  \backgroundpath{
    \southwest \pgf@xa=\pgf@x \pgf@ya=\pgf@y
    \northeast \pgf@xb=\pgf@x \pgf@yb=\pgf@y
    \pgfmathsetlength\pgfutil@tempdima{\pgfkeysvalueof{/pgf/parallelepiped
      offset x}}
    \pgfmathsetlength\pgfutil@tempdimb{\pgfkeysvalueof{/pgf/parallelepiped
      offset y}}
    \def\ppd@offset{\pgfpoint{\pgfutil@tempdima}{\pgfutil@tempdimb}}
    \pgfpathmoveto{\pgfqpoint{\pgf@xa}{\pgf@ya}}
    \pgfpathlineto{\pgfqpoint{\pgf@xb}{\pgf@ya}}
    \pgfpathlineto{\pgfqpoint{\pgf@xb}{\pgf@yb}}
    \pgfpathlineto{\pgfqpoint{\pgf@xa}{\pgf@yb}}
    \pgfpathclose
    \pgfpathmoveto{\pgfqpoint{\pgf@xb}{\pgf@ya}}
    \pgfpathlineto{\pgfpointadd{\pgfpoint{\pgf@xb}{\pgf@ya}}{\ppd@offset}}
    \pgfpathlineto{\pgfpointadd{\pgfpoint{\pgf@xb}{\pgf@yb}}{\ppd@offset}}
    \pgfpathlineto{\pgfpointadd{\pgfpoint{\pgf@xa}{\pgf@yb}}{\ppd@offset}}
    \pgfpathlineto{\pgfqpoint{\pgf@xa}{\pgf@yb}}
    \pgfpathmoveto{\pgfqpoint{\pgf@xb}{\pgf@yb}}
    \pgfpathlineto{\pgfpointadd{\pgfpoint{\pgf@xb}{\pgf@yb}}{\ppd@offset}}
  }
}

\makeatletter
\tikzset{anchor/.append code=\let\tikz@auto@anchor\relax,
  add font/.code=%
    \expandafter\def\expandafter\tikz@textfont\expandafter{\tikz@textfont#1},
  left delimiter/.style 2 args={append after command={\tikz@delimiter{south east}
    {south west}{every delimiter,every left delimiter,#2}{south}{north}{#1}{.}{\pgf@y}}}}
\tikzstyle{sms} = [rectangle callout, draw, very thick, rounded corners, minimum height=20pt]
\makeatletter
\tikzset{anchor/.append code=\let\tikz@auto@anchor\relax,
  add font/.code=%
    \expandafter\def\expandafter\tikz@textfont\expandafter{\tikz@textfont#1},
  left delimiter/.style 2 args={append after command={\tikz@delimiter{south east}
    {south west}{every delimiter,every left delimiter,#2}{south}{north}{#1}{.}{\pgf@y}}}}
\tikzstyle{sms} = [rectangle callout, draw,very thick, rounded corners, minimum height=20pt]
\usetikzlibrary{positioning,calc}

\tikzset{
  mybackground9/.style={execute at end picture={
        \begin{scope}[on background layer]
          \draw[black,fill=Black!5!Sepia!1,rounded corners=6ex] (current bounding box.south west)
                    rectangle (current bounding box.north east);
          \node[draw,fill=white,ellipse,anchor=west,inner sep=1pt,minimum width=4ex] at (current bounding box.north
                   west){#1};
        \end{scope}
    }},
}

\tikzset{
  mybackground10/.style={execute at end picture={
        \begin{scope}[on background layer]
          \draw[black] (current bounding box.south west)
                    rectangle (current bounding box.north east);
          \node[draw,fill=white,ellipse,anchor=west,inner sep=1pt,minimum width=4ex] at (current bounding box.north
                   west){#1};
        \end{scope}
    }},
}

\tikzset{l3 switch/.style={
    parallelepiped,fill=switch, draw=white,
    minimum width=0.75cm,
    minimum height=0.75cm,
    parallelepiped offset x=1.75mm,
    parallelepiped offset y=1.25mm,
    path picture={
      \node[fill=white,
        circle,
        minimum size=6pt,
        inner sep=0pt,
        append after command={
          \pgfextra{
            \foreach \angle in {0,45,...,360}
            \draw[-latex,fill=white] (\tikzlastnode.\angle)--++(\angle:2.25mm);
          }
        }
      ]
       at ([xshift=-0.75mm,yshift=-0.5mm]path picture bounding box.center){};
    }
  },
  ports/.style={
    line width=0.3pt,
    top color=gray!20,
    bottom color=gray!80
  },
  rack switch/.style={
    parallelepiped,fill=white, draw,
    minimum width=1.25cm,
    minimum height=0.25cm,
    parallelepiped offset x=2mm,
    parallelepiped offset y=1.25mm,
    xscale=-1,
    path picture={
      \draw[top color=gray!5,bottom color=gray!40]
      (path picture bounding box.south west) rectangle
      (path picture bounding box.north east);
      \coordinate (A-west) at ([xshift=-0.2cm]path picture bounding box.west);
      \coordinate (A-center) at ($(path picture bounding box.center)!0!(path
        picture bounding box.south)$);
      \foreach \x in {0.275,0.525,0.775}{
        \draw[ports]([yshift=-0.05cm]$(A-west)!\x!(A-center)$)
          rectangle +(0.1,0.05);
        \draw[ports]([yshift=-0.125cm]$(A-west)!\x!(A-center)$)
          rectangle +(0.1,0.05);
       }
      \coordinate (A-east) at (path picture bounding box.east);
      \foreach \x in {0.085,0.21,0.335,0.455,0.635,0.755,0.875,1}{
        \draw[ports]([yshift=-0.1125cm]$(A-east)!\x!(A-center)$)
          rectangle +(0.05,0.1);
      }
    }
  },
  server/.style={
    parallelepiped,
    fill=white, draw,
    minimum width=0.35cm,
    minimum height=0.75cm,
    parallelepiped offset x=3mm,
    parallelepiped offset y=2mm,
    xscale=-1,
    path picture={
      \draw[top color=gray!5,bottom color=gray!40]
      (path picture bounding box.south west) rectangle
      (path picture bounding box.north east);
      \coordinate (A-center) at ($(path picture bounding box.center)!0!(path
        picture bounding box.south)$);
      \coordinate (A-west) at ([xshift=-0.575cm]path picture bounding box.west);
      \draw[ports]([yshift=0.1cm]$(A-west)!0!(A-center)$)
        rectangle +(0.2,0.065);
      \draw[ports]([yshift=0.01cm]$(A-west)!0.085!(A-center)$)
        rectangle +(0.15,0.05);
      \fill[black]([yshift=-0.35cm]$(A-west)!-0.1!(A-center)$)
        rectangle +(0.235,0.0175);
      \fill[black]([yshift=-0.385cm]$(A-west)!-0.1!(A-center)$)
        rectangle +(0.235,0.0175);
      \fill[black]([yshift=-0.42cm]$(A-west)!-0.1!(A-center)$)
        rectangle +(0.235,0.0175);
    }
  },
}
\pgfplotsset{compat=1.16}
\usetikzlibrary{calc, shadings, shadows, shapes.arrows}

\tikzset{%
  interface/.style={draw, rectangle, rounded corners, font=\LARGE\sffamily},
  ethernet/.style={interface, fill=yellow!50},
  serial/.style={interface, fill=green!70},
  speed/.style={sloped, anchor=south, font=\large\sffamily},
  route/.style={draw, shape=single arrow, single arrow head extend=4mm,
    minimum height=1.7cm, minimum width=3mm, white, fill=switch!20,
    drop shadow={opacity=.8, fill=switch}, font=\tiny}
}
\makeatletter
\pgfdeclareradialshading[tikz@ball]{cloud}{\pgfpoint{-0.275cm}{0.4cm}}{%
  color(0cm)=(tikz@ball!75!white);
  color(0.1cm)=(tikz@ball!85!white);
  color(0.2cm)=(tikz@ball!95!white);
  color(0.7cm)=(tikz@ball!89!black);
  color(1cm)=(tikz@ball!75!black)
}
\tikzoption{cloud color}{\pgfutil@colorlet{tikz@ball}{#1}%
  \def\tikz@shading{cloud}\tikz@addmode{\tikz@mode@shadetrue}}
\makeatother

\tikzset{my cloud/.style={
     cloud, draw, aspect=2,
     cloud color={gray!5!white}
  }
}

\usepackage{array}
\newcommand{\thickhline}{%
    \noalign {\ifnum 0=`}\fi \hrule height 1pt
    \futurelet \reserved@a \@xhline
}
\newcolumntype{"}{@{\hskip\tabcolsep\vrule width 1pt\hskip\tabcolsep}}

\IEEEoverridecommandlockouts
\IEEEpubid{\makebox[\columnwidth]{978-3-903176-31-7~\copyright~2020 IFIP\hfill} \hspace{\columnsep}\makebox[\columnwidth]{ }}

\usepackage{etoolbox}
\makeatletter
\patchcmd{\@makecaption}
  {\scshape}
  {}
  {}
  {}
\makeatother

\begin{document}

\title{IT Intrusion Detection Using Statistical Learning and Testbed Measurements}

\author{\IEEEauthorblockN{Xiaoxuan Wang \IEEEauthorrefmark{2} and
 Rolf Stadler\IEEEauthorrefmark{2}}

 \IEEEauthorblockA{\IEEEauthorrefmark{2}
Dept. of Computer Science, KTH Royal Institute of Technology, Sweden
 }
  \newline
Email: \{xiaoxuan, stadler\}@kth.se
\\
February 20, 2024
}

\maketitle

\thispagestyle{plain}
\pagestyle{plain}

\begin{abstract}
\label{sec:abstract}
We study automated intrusion detection in an IT infrastructure, specifically the problem of identifying the start of an attack, the type of attack, and the sequence of actions an attacker takes, based on continuous measurements from the infrastructure. 
We apply statistical learning methods, including Hidden Markov Model (HMM), Long Short-Term Memory (LSTM), and Random Forest Classifier (RFC) to map sequences of observations to sequences of predicted attack actions. 
In contrast to most related research, we have abundant data to train the models and evaluate their predictive power. The data comes from traces we generate on an in-house testbed where we run attacks against an emulated IT infrastructure. 
Central to our work is a machine-learning pipeline that maps measurements from a high-dimensional observation space to a space of low dimensionality or to a small set of observation symbols. Investigating intrusions in offline as well as online scenarios, we find that both HMM and LSTM can be effective in predicting attack start time, attack type, and attack actions. If sufficient training data is available, LSTM achieves higher prediction accuracy than HMM. HMM, on the other hand, requires less computational resources and less training data for effective prediction. Also, we find that the methods we study benefit from data produced by traditional intrusion detection systems like SNORT.     
\end{abstract}

\begin{IEEEkeywords}

automated security, intrusion detection, Hidden Markov Model, Long Short-Term Memory, SNORT, forensics
\end{IEEEkeywords}

\section{Introduction}
\label{sec:introduction}

Intrusion detection systems traditionally rely on a static set of rules for discovering possible attacks. The rules are configured by domain experts, which makes the approach flexible but expensive to maintain in the face of evolving attacks and changing IT infrastructures \cite{snort,ids_survey}. 

Recent research studies automated detection methods where the rules are encoded in the parameters of statistical models that are learned from observations. In this paper, we study three such methods, the Hidden Markov Model (HMM), the Long Short-Term Memory (LSTM), and the Random Forest Classifier (RFC). They are attractive to researchers because they are well-understood and have shown broad applicability in science, finance, and engineering. 

HMM maps a sequence of observations, such as measurements, to a corresponding sequence of hidden (i.e., not observable) states, such as attack actions. The learned mapping allows us to predict from a time sequence of observations whether or not an attack is occurring. Furthermore, we can predict the likely start time of the attack and the sequence of actions the attacker takes. (We use the term ‘predict’ here with the same meaning as ‘estimate’ as is custom in statistical learning.)

Similarly, LSTM can map a sequence of measurements to the corresponding attack action sequence and therefore can be used to make the above-mentioned predictions. RFC, on the other hand, is not designed for sequence-to-sequence learning, but can only map a single measurement to a single action. While RFC can make the same type of predictions, it does not take into account the time dependencies between actions.

The use case we consider in this work is an attack on the IT infrastructure of an organization, whereby an attacker intrudes through a public gateway and attempts to compromise servers of the infrastructure (see Fig. \ref{fig:IT}). We take the perspective of the defender who continuously monitors the infrastructure and receives alerts from a traditional intrusion detection system (IDS). Using trained models and a time series of observations, the defender makes predictions about the presence of an attacker, the type of attack, and the attack actions.

The data to train the models come from traces collected on a testbed we have built at KTH \cite{hammar2023learning}. It emulates the infrastructure depicted in Fig. \ref{fig:IT} and allows us to run various attack scenarios during which we collect measurements and alerts. 

A key issue we address in this work is the high dimensionality of the observation space. There are thousands of possible measurements, statistics, and event counters available on an IT infrastructure, which can be analyzed for suspected attacker actions. In this work, we demonstrate how to reduce the dimensionality of the observation space to a small number, without sacrificing prediction accuracy, which significantly reduces the computational and communication overhead. 


While the use of HMM, LSTM, and RFC for intrusion detection has been proposed before, this paper makes several unique contributions.  First, we provide a systematic, comparative study and evaluation of these three statistical learning methods for the purpose of intrusion detection. We focus on detecting the intrusion start time, the attack type, and the sequence of actions by an attacker.  We study offline and online prediction, as well as prediction with unlabeled data, i.e., with training data that does not include the ground truth about attacker actions. 
Second, we perform training and evaluation of the models using a large number of traces, which we produce on an in-house testbed. The testbed provides a virtualization layer and a time-controlled emulation platform where we run attack scenarios for a realistic use case.
Third, we introduce a machine-learning (ML) pipeline that maps high-dimensional observation vectors into vectors of low dimensionality (or, alternatively, into a small number of observation symbols in the case of HMM), which allows for efficient training in a realistic scenario. 


\section{Use Case}
\label{sec:use_case}
\begin{figure}
  \centering
    \scalebox{0.6}{
      \includegraphics{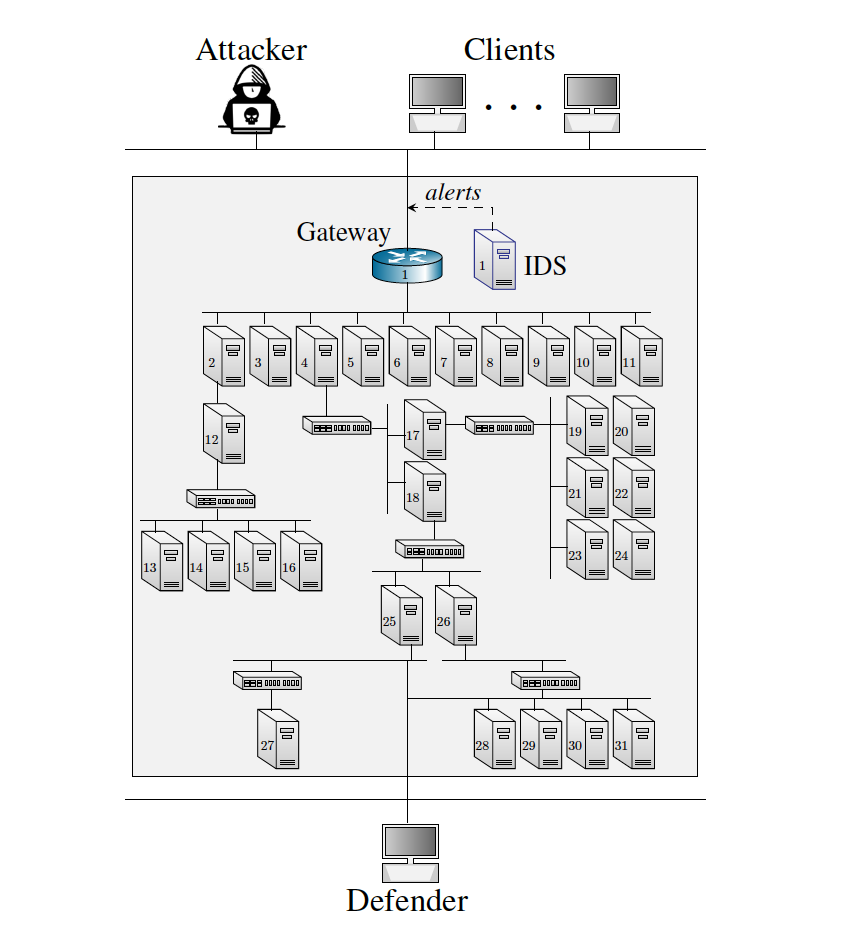}
    }
    \caption{The IT infrastructure in the intrusion detection use case.}
    \label{fig:IT}
  \end{figure}

We consider an intrusion detection use case that involves the IT infrastructure of an organization as depicted in \mbox{Fig. \ref{fig:IT}} \cite{hammar2022learning}. The infrastructure includes a set of servers that run application services and an IDS that logs events in real-time. The operator of this infrastructure, which we call the defender, provides services to a client population and monitors the infrastructure and the services. The clients access the services through a public gateway, which is also open to an attacker. The attacker intrudes on the infrastructure and performs an attack, for instance, retrieving information, installing malicious software, etc. During an attack, the attacker executes a sequence of attack actions including network scans, vulnerability exploits, and tool installation. The defender continuously monitors the infrastructure and analyzes the IDS statistics and measurements from network components and servers. Note that the defender does not directly observe the attacker but only indirectly learns about an attack through the monitoring data, which originates from the actions of the attacker and the clients. Using this data, the defender has the following objectives:

\begin{enumerate}
    \item Predict whether there is an ongoing attack, and if true, predict the start time of the attack.
    \item Predict the type of the attack.
    \item Predict the sequence of attack actions.
\end{enumerate}

The defender can pursue these objectives for offline prediction, where an entire trace of observations is available, which facilitates a forensic investigation. Alternatively, the defender can perform online prediction, where a sequence of observations up to the current time is available, which allows for real-time intrusion prevention.

\section{Problem formulation and approach}
\label{sec:problem_formulation}

We study the use case and model its evolution as a discrete-time dynamical system. At each time step, the attacker executes an action from a finite action set. Depending on the type of attack, the attacker takes a specific sequence of attack actions. The defender, at each time step, collects and processes a vector of observations, which include infrastructure measurements and event statistics. Fundamentally, the problem we want to solve is to find an accurate mapping from the sequence of observations to the sequence of attack actions (which are hidden from the defender). 

In the case of HMM, we interpret a sequence of $T$ attack actions as the hidden state sequence of length $T$ in a HMM. Similarly, we interpret the corresponding sequence of $T$ observations as the sequence of observations in the same model. Given this interpretation, we understand the evolution of the system, i.e., the evolution of an attack in our use case, as the evolution of the HMM. 

The properties of HMM have been thoroughly studied in the literature and many algorithms for HMM have been developed (see Section \ref{sec:theoretical_background}). One of them is the well-known Viterbi algorithm \cite{forney1973viterbi}, which maps a given observation sequence to the most likely sequence of hidden states. 

In order to apply the Viterbi algorithm, we must find the defining parameters of the HMM, for which there are two principal ways. If we have access to sample sequences of hidden states and corresponding observations, we can estimate the model parameters using a supervised learning method. On the other hand, if we have access to sample sequences of observations only (and not to hidden states), we can estimate the model parameters through unsupervised learning. For this work, we obtain sample sequences from traces collected on our testbed, which emulates the infrastructure depicted in \mbox{Fig. \ref{fig:IT}}, where we run different types of attacks (see Section \ref{sec:emulation_collection}).

HMM is usually used with the first-order assumption, which says that the hidden state at time $t$ only depends on the hidden state at time $t-1$ and not on earlier states. In our scenario, this means that an attack action only depends on the action of the previous time step but not on earlier time steps. This is obviously an approximation of reality, and an experimental evaluation will show whether the assumption is justified.

LSTM is a sequence-to-sequence learning method that maps an input sequence to an output sequence. In this work, we use this method to map a sequence of $T$ observations to a sequence of $T$ attack actions. RFC maps an input vector to a class label. In this case, we use the method to map a single observation to a single action. By repeatedly applying RFC, we map a sequence of observations to a sequence of attack actions.

We use the categorical version of a HMM, which assumes a finite set of possible observation symbols. This assumption is made in most of the related research that uses HMM \cite{srivastava2008credit,arnes2006using,zhang2014application,shawly2019architectures,ghafir2019hidden,holgado2017real}. Since the number of model parameters increases linearly with the number of possible observations, we attempt to keep the number small in order to limit the sample size required to accurately estimate the model and to reduce the model complexity. In our use case, the observation vector is high-dimensional: it contains 50 observation attributes, such as host metrics and IDS alert counts. Consequently, we must find a mapping from the high-dimensional observation space onto a small set of observation symbols, which are used in the HMM. In this work, we apply feature selection and clustering techniques to accomplish this mapping (see Section \ref{sec:data_preprocessing}). 

For LSTM and RFC, we do not use observation symbols but numerical values. As the computational complexity of both LSTM and RFC increase linearly with the number of observation attributes, i.e., the dimensionality of the observation space, we aim at a small attribute set that is selected during the data preprocessing stage.

Recall that an attacker performs a sequence of actions from a finite action set. The \emph{`Continue'} action is a special action in this set. If the attacker performs \emph{`Continue'} at time $t$, it rests during time step $t$ and does not produce any observable change or event. An attack episode always starts with a series of \emph{`Continue'} actions. At time $t_{start}$, the attacker switches to a different action and continues with such actions until the end of the attack. This way of structuring an attack episode keeps the defender guessing about the start time of the attack.

We use the notations described in Table \ref{tab:notations}. Given a trained model and an observation sequence, the objectives of the defender can be formally stated as:
\begin{enumerate}
    \item Predict the start time $t_{start}$ of the attack.
    \item Predict the attack type.
    \item Predict the most likely hidden state sequence $(\widehat{s}_{1},\widehat{s}_{2},...,\widehat{s}_{T})$.
\end{enumerate}

\begin{table}[ht]
    \centering
    \caption{Table of notation}
    \label{notation}
    \scalebox{0.9}{
    \begin{tabular}{|c|c|}
    \hline 
    $D$& number of observation attributes\\
    \hline
    $N$& number of hidden states\\
    \hline 
    $M$& number of observation symbols\\
    \hline 
    $T$& sequence length\\
    \hline 
    $\lambda$& statistical model\\
    \hline 
    $Q=\left \{ q_{1},q_{2},...,q_{N}\right \}$& set of (hidden) states\\
    \hline
    $V=\left \{ v_{1},v_{2},...,v_{M}\right \}$& set of observations\\
    \hline
    $S=( s_{1},s_{2},...,s_{T})$& (hidden) state sequence\\
    \hline
    $O=( o_{1},o_{2},...,o_{T})$& observation sequence\\
    \hline
    \end{tabular}}
    \label{tab:notations}
\end{table}

Throughout this paper, we are using the terms `attack' and  `intrusion' as synonyms. We find both terms in the literature with the same meaning.

\begin{figure}
  \centering
    \scalebox{0.16}{
      \includegraphics{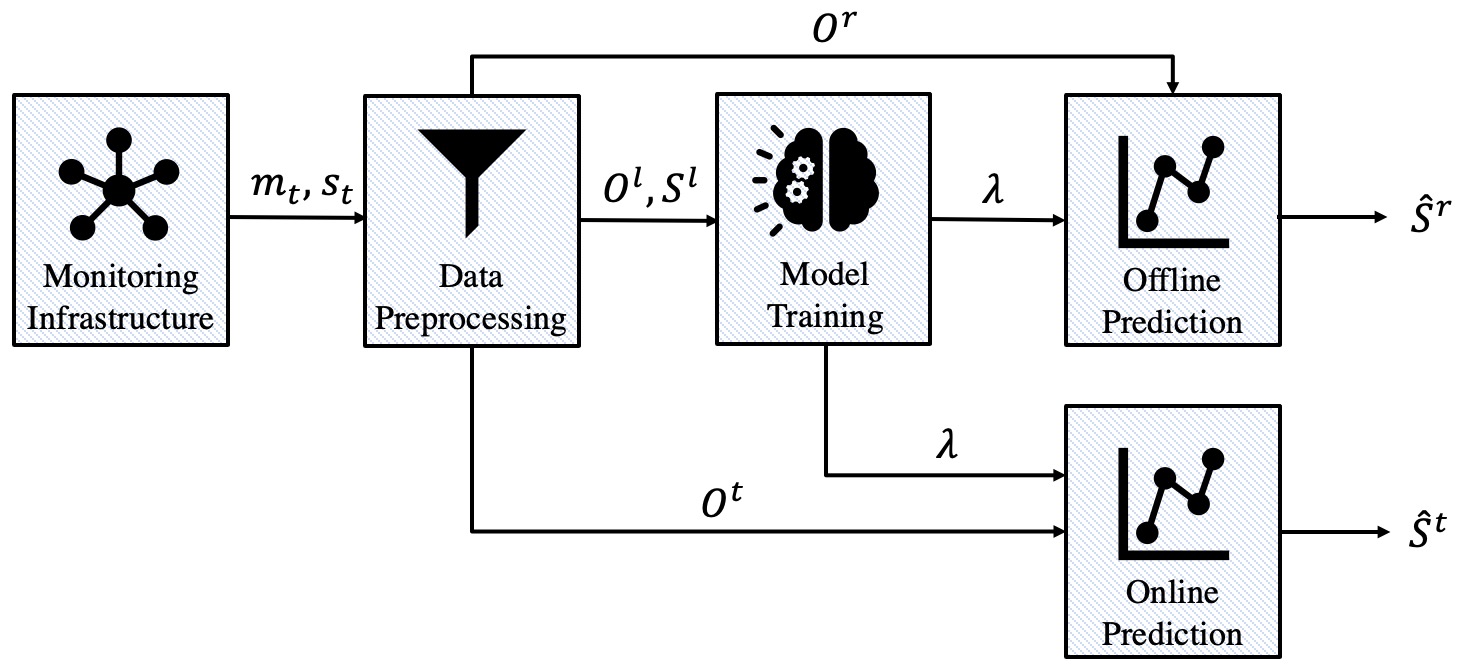}
    }
    \caption{ML pipeline of the proposed approach. Models we consider are HMM, LSTM, and RFC.}
    \label{fig:pipeline}
  \end{figure}

Fig. \ref{fig:pipeline} shows our approach to intrusion detection using statistical models, which we illustrate with a ML pipeline. The monitoring system on the left side of the figure regularly collects an observation vector $m_{t}$ together with the attack action $s_{t}$. The preprocessing step maps the high-dimensional observation $m_{t}$ onto a low-dimensional observation vector or an observation symbol. During the training phase, sample sequences of observations $O^{l}$ and attack actions $S^{l}$ are used to estimate the parameters for statistical model $\lambda$. (In the case of unsupervised learning of HMM, only sample sequences of observations $O^{l}$ are used.) During the phase of online prediction, the current sequence of observations $O^{t}$ is mapped onto a sequence of predicted attack actions $\widehat{S}^{t}$ using the model $\lambda$. 

\section{Methods}
\label{sec:theoretical_background}

\subsection{Hidden Markov Model (HMM)}
\begin{figure}
  \centering
    \scalebox{0.25}{
      \includegraphics{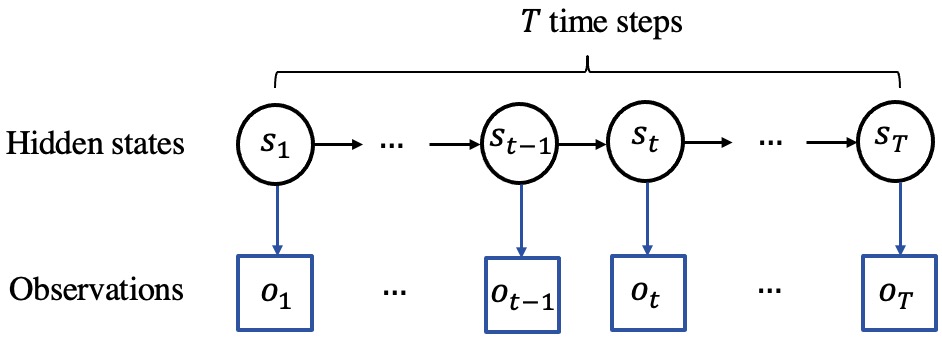}
    }
    \caption{Structure of Hidden Markov Model.}
    \label{fig:hmm}
  \end{figure}

A HMM is a statistical model that describes a Markov process with hidden states and associated observations. It captures the probabilistic relationship between a hidden state and an observation, as shown in Fig. \ref{fig:hmm}. A HMM $\lambda$ can be defined as the triple \cite{rabiner1989tutorial}:
\begin{equation}
    \lambda =(A,B,\pi )
\end{equation}
$A$ is the transition matrix between hidden states:
\begin{equation}
    A=\left \{ a_{ij} \right \}, a_{ij}=P(s_{t+1}=q_{j}|s_{t}=q_{i}), 1\leq i,j\leq  N
\end{equation}
$B$ is the matrix that gives the probability of state $q_{j}$ producing the observation $v_{k}$:
\begin{equation}
\label{equ:b}
    B=\left \{ b_{j}(k) \right \}, b_{j}(k)=P(o_{t}=v_{k}|s_{t}=q_{j}), 1\leq k\leq  M
\end{equation}
$\pi$ gives the probability of the initial state at time $t=1$:
\begin{equation}
\label{equ:pi}
    \pi =\left \{ \pi _{i} \right \}, \pi _{i}=P(s_{1}=q_{i}), 1\leq i\leq N
\end{equation}
$a_{ij}$, $ b_{j}(k)$, $\pi _{i}$ are probabilities with values in the interval $[0,1]$. Moreover:
\begin{equation}
    \sum\limits_{i=1}^{N}\pi _{i}=\sum\limits_{j=1}^{N}a _{ij}=\sum\limits_{k=1}^{M}b _{j}(k)=1
\end{equation}
It is customary to make the first-order assumption for the HMM $\lambda$:
\begin{enumerate}
    \item $P(s_{t}|s_{1},s_{2},...,s_{t-1})=P(s_{t}|s_{t-1})$
    \item $P(o_{t}|s_{1},...,s_{t},o_{1},...,o_{t-1})=P(o_{t}|s_{t})$
\end{enumerate}

Given a HMM $\lambda$ and an observation sequence $O=( o_{1},o_{2},...,o_{T})$, we calculate the probability of the observation sequence being generated by $\lambda$:
\begin{equation}
P(O|\lambda)=\sum\limits_{i=1}^{N}P(o_{1},o_{2},...,o_{T},s_{T}=q_{i})=\sum\limits_{i=1}^{N}\alpha _{T}(i) 
\end{equation}
where $\alpha _{t}(i)$ is the forward variable which represents the probability that the hidden state is $q_{i}$ when observing the sequence $(o_{1},o_{2},...,o_{t})$.
\begin{equation}
\label{equ:alpha}
    \alpha _{t}(i)=P(o_{1},o_{2},...,o_{t},s_{t}=q_{i}|\lambda )
\end{equation}

Given a HMM $\lambda$ and an observation sequence $O=( o_{1},o_{2},...,o_{T})$, we want to find the most likely sequence $\widehat{S}=(\widehat{s}_{1},\widehat{s}_{2},...,\widehat{s}_{T})$. This problem can be solved with the Viterbi algorithm, a dynamic programming algorithm. The computational complexity of the Viterbi algorithm is $O(N^{2}T)$.

Training a HMM $\lambda$ means estimating its parameters $(A,B,\pi)$. When sample sequences of hidden states and related sequences of observations are available, the parameters can be estimated by counting occurrences, which constitutes a supervised learning method:
\begin{equation}
    \widehat{a}_{ij}=\frac{A_{ij}}{\sum\limits_{j=1}^{N}A_{ij}},\ A_{ij}\ is\ the\ count\ of\ transitions\ q_{i} \to q_{j}
\end{equation}
\begin{equation}
    \widehat{b}_{j}(k)=\frac{B_{jk}}{\sum\limits_{k=1}^{M}B_{jk}},\ B_{jk}\ is\ the\ count\ of\ v_{k}\ observed\ in\ q_{j}
\end{equation}
\begin{equation}
    \widehat{\pi }_{i}=\frac{\Pi _{i}}{\sum\limits_{i=1}^{N}\Pi _{i}},\ \Pi _{i}\ is\ the\ count\ of\ S\ starts\ with\ q_{i}
\end{equation}

In case only sample sequences of observations are available, the Baum-Welch algorithm \cite{rabiner1986introduction} allows for unsupervised training. It is an expectation-maximization algorithm that attempts to maximize $P(O|\lambda )$. The computational complexity of the Baum-Welch algorithm is $O(N^{2}TIL)$ where $I$ is the number of iterations and $L$ is the number of training sequences.

\subsection{Long Short-Term Memory (LSTM)}

LSTM \cite{hochreiter1997long} has a specific recurrent neural network architecture that is capable of learning long-term dependencies. For this work, we choose a LSTM architecture with the length of the input sequence equal to the length of the output sequence. The input is a sequence of measurements and the output is a sequence of attack actions. While input values are numerical, output values are categorical.

The computational complexity of training a LSTM is $O(TH^2+THD)$ where $H$ is the number of cells in hidden layers \cite{rotman2021shuffling}. We use the \textbf{tensorflow.keras} package \cite{LSTM} for training and evaluation. 


\subsection{Random Forest Classifier (RFC)}
RFC \cite{breiman2001random} is a supervised ensemble learning method that aggregates the results of multiple decision trees to determine the output. In this work, we not only use RFC for predicting attack actions but also in the data preprocessing phase to reduce the number of attributes for model training. 


The computational complexity of RFC is $O(DFR\log R)$ where $F$ is the number of trees, and $R$ is the number of samples \cite{TBcomplexity}. We use the \textbf{sklearn.ensemble.RandomForestClassifier} package \cite{RFclassifier} with default parameters.

\section{Emulating attacks and collecting measurements}
\label{sec:emulation_collection}

We have built a testbed at KTH which emulates the infrastructure and the configuration shown in Fig. \ref{fig:IT}. On the testbed, we emulate physical hosts, network connections, the client population, as well as the attacker and the defender. For this research, we have executed thousands of attack episodes and have obtained traces from the measurements and events observed by the defender. The length of a time step during these experiments has been 30 seconds. More information about the emulation platform is available at \cite{hammar2023learning}.

Running an attack experiment on the emulation platform involves running a series of attack episodes. Each episode starts at time $t=1$, and the intrusion starts at a random time drawn from a geometric distribution $t_{start} \sim Ge(p=0.2)$. We execute two types of attacks, which are listed in \mbox{Table \ref{tab:attack_actions}}. Both rely on six different attack actions. For this work, we collected data from 2000 attack episodes, 1000 for each attack type.

\begin{table}
\centering
\resizebox{0.48\textwidth}{!}{
\begin{tabular}{lll} \toprule
  {\textit{Attack steps}} & {\textbf{Attack type 1}} & {\textbf{Attack type 2}}\\ \midrule
  1 & Ping Scan & Ping Scan\\
  2 & CVE-2017-7494 exploit & Install Tools\\
  3 & Network Service Login & Network Service Login\\
  4 & Install Tools & Install Tools\\
  5 & Ping Scan & Network Service Login\\
  6 & DVWA SQL Injection Exploit \cite{dvwa} & DVWA SQL Injection Exploit\\
  7 & Network Service Login & CVE-2017-7494 exploit\\
  8 & Install Tools & Network Service Login\\
  9 & Ping Scan & CVE-2017-7494 exploit\\
  10 & CVE-2015-1427 Exploit & Ping Scan\\
  11 & Network Service Login & Ping Scan\\
  12 & Install Tools & Install Tools\\
  13 & Ping Scan & Network Service Login\\
  14 & CVE-2017-7494 exploit & Ping Scan\\
  15 & Network Service Login & CVE-2015-1427 Exploit\\
  16 & Install Tools & Ping Scan\\
  17 & Ping Scan & Install Tools\\
  \bottomrule\\
\end{tabular}
}
\caption{Two attack types with attack action sequences. Actions that exploit vulnerabilities in specific software products are identified by the vulnerability identifiers in the Common Vulnerabilities and Exposures (CVE) database \cite{cve}.}
\label{tab:attack_actions}
\end{table}

\begin{figure}
  \centering
    \scalebox{0.24}{
      \includegraphics{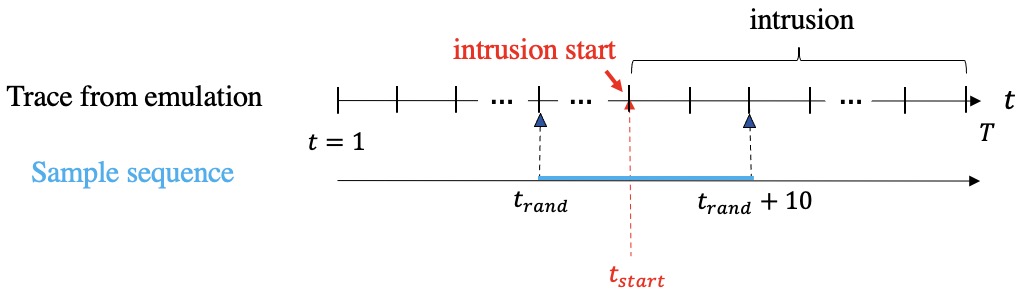}
    }
    \caption{Generating a sample sequence from a testbed trace.}
    \label{fig:preprocess}
  \end{figure}

From a trace collected on the emulation platform, we create a sample sequence for training the models as illustrated in \mbox{Fig. \ref{fig:preprocess}}. More precisely, we obtain two sequences per episode: a sequence of states, i.e., attack actions, and a corresponding sequence of observations. 

We choose the length of a sample sequence to be 10, and we choose the start time of the sample sequence $t_{rand}$ uniformly at random in the interval $\left [ t_{start}-9,t_{start} \right ]$, which ensures that each sample sequence includes the intrusion start time. This procedure gives us 1000 sample sequences of states and observations for each attack type to train and evaluate the models.

\section{Data preprocessing}
\label{sec:data_preprocessing}

We map high-dimensional observation vectors to low-dimensional ones. For HMM, we further map the low-dimensional observation vectors to a small set of observation symbols. As we argued in Section \ref{sec:problem_formulation}, this process reduces the computational complexity of all three methods and thus the training time and sample size needed for an accurate model. In addition, it can reduce the monitoring overhead for online prediction. 

Our method consists of the following three steps. The first two steps apply to all three methods, and the last step only applies to HMM.

\begin{enumerate}
    \item We remove redundant observation attributes that do not affect the accuracy of the prediction tasks. First, we remove attributes whose values remain constant across time. Second, we compute the pairwise correlation between attributes. When an attribute pair shows a high correlation (larger than 0.9), we remove one of the attributes. 
    \item We rank the remaining observation attributes and use the top-ranked ones. To achieve this, we train a RFC model that predicts from a measurement vector whether there is an ongoing attack. From the training process, we obtain the ranking of the attributes.
    \item We consider the set of observation vectors with the reduced set of attributes and partition the reduced observation space using a clustering algorithm. Each partition obtained through this process is then associated with an observation symbol. The clustering algorithm we use is based on a Gaussian mixture model whereby the number of components in the model equals the number of observation symbols.
\end{enumerate}

The traces collected from running attack scenarios on our testbed contain observation vectors with 50 attributes. Interestingly, more than half of them turn out to have constant values and are therefore redundant. We apply the correlation analysis to the remaining 14 attributes, which leads us to remove the additional three attributes. Training the RFC produces the ranking of the remaining 11 attributes. Four of them are alert counts from the IDS, which is a SNORT system \cite{snort}. The rest are statistics from servers. 

Since the two attack types we are considering in this work are based on six different attack actions (see Table \ref{tab:attack_actions}), we choose the number of observation symbols in the HMM to be six. We performed evaluations with larger numbers but found that those cases did not result in increasing the prediction accuracy.

\section{Evaluation results}
\label{sec:learning_hmm_evaluation}

\subsection{Evaluation metrics}
We define four metrics for evaluating the prediction accuracy of the trained models. $l$ stands for the number of sample sequences for testing. Variables with a hat symbol denote predictions. For example, $\widehat{s}_{t}^{i}$ stands for the prediction of the ground truth $s_{t} ^{i}$.

\begin{itemize}
    \item \textbf{$Acc_{start}$}: the accuracy of correctly predicting the intrusion start time.
    \begin{equation}
    Acc_{start}=\frac{1}{l}\sum\limits_{i=1}^{l}\mathbbm{1}\left \{ t_{start} ^{i}== \widehat{t}_{start}^{i} \right \}
    \end{equation}
    \item \textbf{$Acc_{type}$}: the fraction of correctly classified observation sequences. $c$ represents the attack type.
    \begin{equation}
    Acc_{type}=\frac{1}{l}\sum\limits_{i=1}^{l}\mathbbm{1}\left \{ c^{i}== \widehat{c}^{i} \right \}
    \end{equation}
    \item \textbf{$Acc_{action}$}: the fraction of correctly predicted single actions. 
    \begin{equation}
    Acc_{action}=\frac{1}{lT}\sum\limits_{i=1}^{l}\sum\limits_{t=1}^{T}\mathbbm{1}\left \{ s_{t} ^{i}== \widehat{s}_{t}^{i} \right \}
    \end{equation}
    \item \textbf{$Acc_{sequence}$}: the accuracy of correctly predicting the entire sequence of actions. 
    \begin{equation}
    Acc_{sequence}=\frac{1}{l}\sum\limits_{i=1}^{l}\mathbbm{1}\left \{ S^{i}== \widehat{S}^{i} \right \}
    \end{equation}  
\end{itemize}

The values of all four metrics lie between 0 and 1. A higher value indicates a higher accuracy of the trained model. The metrics can be interpreted as a probability of accurately predicting the intrusion start time, the type of attack, the current attack action, or the sequence of attack actions, respectively.

\subsection{Model training and evaluation}

To train statistical models, we create 2000 sequence pairs, 1000 for each attack type (see Section \ref{sec:emulation_collection}). For each attack type, we then divide, uniformly at random, the sequence pairs into two sets whereby we put 70\% into the set to train a model and 30\% into the set to evaluate the prediction accuracy of the trained model. 

As explained in Section \ref{sec:theoretical_background}, we can obtain a HMM through supervised learning by using both observation sequences and state sequences. Alternatively, we can obtain a HMM through unsupervised learning by using only observation sequences. Note that a HMM trained through unsupervised learning can not be used to directly predict attack actions. Instead, it allows only to predict state labels whereby each label has a one-to-one correspondence with an attack action. The (most probable) mapping from labels to the actions can be learned when sample pairs of observation and corresponding state sequences are available. For the evaluation in this paper, we take 100 such pairs from the training set and compute $Acc_{avg}$ for all possible mappings from state labels to attack actions. (In our case, there are 6! = 720 such mappings.) Then we use the mapping with the highest $Acc_{action}$ value for mapping a state label sequence to an attack action sequence.

LSTM is trained using both observation sequences and state sequences. RFC takes (observation, attack action) pairs instead of the sequence pairs as training samples. For evaluation, the classifier predicts an attack action at each time step of the observation sequence and these predictions constitute the final predicted sequence.

We perform model training and evaluation for two different subsets of observation attributes. Recall that the method for producing observation symbols discussed in Section \ref{sec:data_preprocessing} reduces the number of observation attributes to 11 for the trace from our scenario. Extensive evaluation not further detailed here shows that only a small number of those attributes are needed for accurate prediction and using a full observation space leads to worse performance. In the following, we present evaluation results for two sets of attributes. The first set contains only the top-ranked attribute `\emph{misc-activity}' which is a SNORT alert count. The second set contains the top four attributes, three of which are SNORT alert counts. 

Given an observation sequence, we predict the attack type using HMM in the following way. First, we train a separate HMM for each attack type. We compute the probability of the occurrence of the observation sequence for each HMM using the forward variable (see Section \ref{sec:emulation_collection}). The HMM that gives the highest probability identifies the attack type. When using LSTM or RFC, we compare the predicted attack action sequence to the sequence listed in Table \ref{tab:attack_actions} and the similarity value suggests the attack type.

\subsection{Results for offline prediction (forensics)}
\label{subsec:evaluation}
Each value in Table \ref{tab:all_combine_1} is the average of 10 runs with different random seeds. A run includes creating a training set and a test set as described above, training the models, and evaluating their accuracy.

\begin{table*}[ht]
    \centering
    \caption{Offline training, offline prediction: accuracy for different prediction methods. The results are for Attack 1.}
    \label{tab:all_combine_1}
    \begin{tabular}{|c|c|c|c|c|c|c|c|c|}
    \hline
     \multirow{2}{*}{Method} & \multicolumn{4}{|c|}{Top observation attribute}& \multicolumn{4}{|c|}{Top 4 observation attributes}\\
    \cline{2-9}
     ~ & $Acc_{start}$ & $Acc_{type}$ & $Acc_{action}$ & $Acc_{sequence}$ & $Acc_{start}$ & $Acc_{type}$ & $Acc_{action}$ & $Acc_{sequence}$\\
    \hline 
    Unsupervised HMM & 0.993 & 0.993 & 0.779 & 0.058 & 0.993 & 1 & 0.689 & 0.046\\
    \hline 
    Supervised HMM & 0.995 & 0.995 & 0.919 & 0.215 & 0.994 & 0.998 & 0.924 & 0.388\\
    \hline 
    LSTM & 0.957 & 0.970 & 0.980 & 0.917 & 0.978 & 0.985 & 0.990 & 0.967\\
    \hline 
    RFC & 0.957 & 0.782 & 0.463 & 0 & 0.993 & 0.763 & 0.459 & 0\\
    \hline
    \end{tabular}
\end{table*}

The left part of Table \ref{tab:all_combine_1} shows the prediction accuracy for the different prediction methods we study and for the case where we choose the attribute set with the top-ranked attribute from the data preprocessing phase. The right part of Table \ref{tab:all_combine_1} contains the same information for the case where we choose the attribute set with the four top-ranked attributes. The figures in the table refer to Attack 1. The corresponding figures for Attack 2 are very similar. The table allows us to answer the questions we pose in Sections \ref{sec:use_case} and \ref{sec:problem_formulation}. 

We observe an accuracy above 0.95 for the prediction of the 
\textbf{\emph{intrusion start time}} in all cases, suggesting that all three methods would predict the intrusion start time correctly in 95\% of the cases on unseen data. Similarly, we notice a high accuracy for the \textbf{\emph{attack type}} classification with values above 0.97 when using HMM or LSTM. Regarding the accuracy of predicting \textbf{\emph{attack actions}}, the values are significantly lower, ranging from 0.45 to 0.99. It turns out that predicting the \textbf{\emph{sequence of actions}} is hard when using HMM or RFC. The values for using LSTM are much higher and are above 0.91. 

When comparing the different methods, we observe that for the prediction of the intrusion start time, HMM, RFC, and LSTM have similar performance. For the prediction of attack actions, LSTM performs the best and RFC performs the worst. When comparing the results for supervised HMM with unsupervised HMM, we find a similar performance for intrusion start time prediction and attack type classification. However, supervised HMM clearly outperforms unsupervised HMM for attack action prediction.

Regarding the two different observation attribute sets, the prediction accuracy values tend to be better if the larger attribute set is used.

\subsection{Results for online prediction (real-time supervision)}

While all models are trained offline, they can be used for real-time prediction once a full sequence of observations is available, and we expect they show the prediction accuracy given in Table \ref{tab:all_combine_1}. If a complete sequence of observation is not available, for instance after the start of a system, we need an online method for real-time prediction. 

As HMM can process sequences of different lengths, it is applicable to an online prediction scenario where we collect observations at each time step and make a new prediction at the end of the time step. We use supervised HMM with an offline-trained model to predict the current attack action in an online fashion. 

\begin{figure}[ht]
  \centering
    \scalebox{0.5}{
      \includegraphics{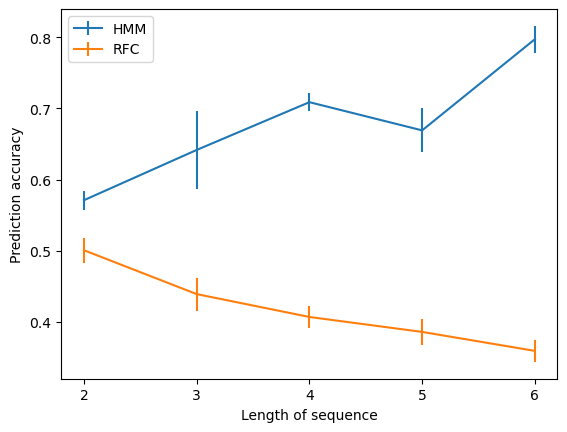}
    }
    \caption{Offline training, online prediction: prediction accuracy of the attack action at the current time step in function of the observation sequence length. The results are for Attack 1.}
    \label{fig:acc_now}
  \end{figure}

Fig. \ref{fig:acc_now} shows the prediction accuracy of the attack action at the current time in function of the length of the observation sequence. Each data point in the figure represents the mean value from 10 runs with different random seeds, and the bar shows the standard deviation. The measurements are for Attack 1 and the attribute set with the top attribute. For HMM, we observe that the prediction accuracy increases with the sequence length, as we expect. For instance, for the observation sequence of length six, the attack action can be predicted with an accuracy of 0.8. There is an aberration where the accuracy decreases when the sequence length grows from four to five. We can not fully explain this behavior. 

LSTM is less suitable for online prediction since it requires a specific model for each sequence length. In contrast, RFC takes input of length one and thus is applicable to online prediction. Fig. \ref{fig:acc_now} shows that the prediction accuracy for RFC is much lower than that for HMM. We also observe that the accuracy is not constant across sequence lengths, which suggests that, for a given attacker action, the following action is not randomly chosen among all possible actions.

\subsection{Discussion of the evaluation results}

The experimental evaluation gives us the following insights:

\begin{enumerate}
    \item If we are only interested in \textbf{\emph{predicting the intrusion start time, RFC is the best choice}}, because it achieves similar accuracy as HMM and LSTM at lower computational complexity.
    \item \textbf{\emph{LSTM generally allows for better prediction than HMM and RFC}}. We explain this by the fact that LSTM can capture long-term dependencies among attack actions, whereas HMM follows the first-order assumption that an attack action only depends on the action in the previous time step, not on any action before that. The drawback of LSTM is that it has a higher complexity than HMM and  RFC, and it can not be trained with unlabelled data. This makes LSTM unpractical when the amount of training data is insufficient or attack action sequences are not available. 
    \item The evaluation figures for HMM show that the first-order assumption can be made in our use case to achieve accurate prediction.
    \item \textbf{\emph{HMM and LSTM outperform RFC in terms of predicting the attack actions}}. This can be explained by the fact that RFC does not capture the temporal information of sequences.
    \item Supervised HMM generally leads to higher prediction accuracy than unsupervised HMM, because supervised HMM is trained with additional labeled training data (i.e., attack action sequences). 
    \item We are surprised that \textbf{\emph{small sets of observation attributes can lead to accurate predictions}}. Note that many of these attributes are alert counts from SNORT.
\end{enumerate}

Overall, we find that \textbf{\emph{HMM is an attractive choice for intrusion detection}}. In our evaluation, it provides predictions with the same accuracy as LSTM at lower computational complexity for training models. In our experiments, we can predict the intrusion start time with an accuracy above 0.99, the attack type above 0.99, and the average attack actions above 0.68. In contrast to LSTM, HMM allows for prediction even if little or no labeled data is available. In addition, HMM supports online prediction with a single model that processes sequences of different lengths. 

We applied HMM and RFC to the dataset CSE-CIC-IDS2018 \cite{sharafaldin2018toward} and obtained a prediction accuracy for the intrusion start time that is very close to the one reported for our traces. (This dataset does not allow for other predictions that we performed in our evaluation.) 

We find several works in the literature that use HMM or LSTM to predict the stage of an attack (See Section \ref{sec:related_work}). While a detailed comparison with our results is difficult due to the difference in the observation data, we find that most results in the literature are consistent with our evaluation.

\section{Related work}
\label{sec:related_work}

When reviewing works that apply HMM to address problems in IT security, we found two directions. First, HMM is used to profile an actor \cite{srivastava2008credit,annachhatre2015hidden,zhao2017software,wang2004modeling,sperotto2009hidden}, for example, an attacker or a normal user in the network environment. Based on the profile, the actor can be identified based on a sequence of observations. Second, HMM is used to identify different stages or states of an attack \cite{arnes2006using,zhang2014application,shawly2019architectures,ghafir2019hidden,holgado2017real,zan2009hidden,chen2016anomaly}. In this work, we follow both directions by constructing HMM that allows us to profile attack types and by training HMM that allows us to predict intrusion start time and attack actions.

In \cite{sperotto2009hidden}, the authors argue that attack-labeled flow data sets are rare for evaluating flow-based intrusion detection techniques. Therefore, they propose a generative method for generating synthetic flow-based time series data under SSH brute-force attacks. The method is based on HMM where the hidden states correspond to the attack phases and attacker activity while the observations correspond to the number of flows, the number of packets, and the number of bytes. HMM is first trained in a supervised fashion with real network data and then the trained model is used to generate synthetic data. 

In \cite{shawly2019architectures}, the authors study an intrusion detection scenario where two different attackers simultaneously intrude on the system. They develop two different methods to predict the attack type and attack stage using HMM. Both methods rely on two HMMs modeling the two attacks, and one of them introduces a special observation symbol that describes the case where an observation does not relate to the considered attack type. 

In \cite{holgado2017real}, the authors propose a method for real-time prediction of an attack stage based on HMM. They consider three stages of a DDoS attack. They collect alerts from the SNORT system and map them onto 10 observation symbols based on analysis of CVE reports \cite{cve}. The models are trained offline using the public DARPA2000 dataset LLDDOS1.0 \cite{darpa}. Their evaluation shows that they can achieve similar prediction accuracy as we report in this paper for supervised learning. For unsupervised learning, however, they achieve much lower accuracy.

Some works combine HMM with neural networks \cite{deshmukh2019attacker,aoudni2022cloud,jiang2005novel}. The authors of \cite{deshmukh2019attacker} introduce the Fusion Hidden Markov Model (FHMM) for profiling attacker behavior in computer networks. A FHMM entails a stochastic ensemble of HMMs to capture the diverse behavior of attackers. The authors partition the set of training sequences into least correlated subsets. For each subset, they train a HMM and aggregate the predictions from all HMMs through a neural network. The authors motivate their approach with the need for a method robust to noise. In \cite{aoudni2022cloud} the authors propose an attack detection and classification framework for zero-day attacks. The predictions of the trained HMM are used to train a deep neural network.

In \cite{dass2021} the authors study the method of classifying a family of attacks using a single HMM instead of using one HMM for each attack type. They argue that different attack types often have similar characteristics and thus may share some observation symbols and hidden states. In our work, we have evaluated and compared both approaches (i.e., training a HMM for each attack type vs. training a single HMM for both attack types) and found that they lead to very similar accuracy values for attack classification.

Similar to HMM, the related works that apply LSTM to intrusion detection fall into two categories. First, LSTM is used as a classifier \cite{staudemeyer2015applying,diro2018leveraging,althubiti2018lstm,laghrissi2021intrusion}. Second, like in this paper, LSTM is used for sequence-to-sequence learning \cite{sai2019advance,zhou2021detecting}.

The authors of \cite{staudemeyer2015applying} claim that their paper is the first successful application of LSTM to intrusion detection. They model network traffic as a time series and apply LSTM to classify the connection records into either normal traffic or one of four attack categories. Different LSTM architectures are evaluated. Like \cite{staudemeyer2015applying}, the authors of \cite{laghrissi2021intrusion} apply LSTM to traffic classification. They evaluate different feature reduction techniques applied before LSTM and find that PCA gives the best results.

\cite{zhou2021detecting} applies LSTM to predict a sequence of attack stages. The observations are Snort messages which are mapped onto a vector space. The method uses an encoder-decoder architecture. One LTSM network encodes a sequence of alerts into a latent feature vector and then another LSTM network decodes this vector to a sequence of predicted attack stages. 

RFC has been applied to intrusion detection as a classifier \cite{gupta2016framework,belavagi2016performance,stefanova2017network,da2017detecting,lashkari2017characterization,mcelwee2017active,choubisa2022simple}. In \cite{stefanova2017network}, the authors propose a two-stage classifier for network intrusion detection. The first stage uses RFC to separate the traffic into the classes “normal” and “attack.” The second stage uses a different classifier to further separate the attack traffic into attack types. In \cite{da2017detecting} RFC is used to detect mobile botnets by classifying the system calls invoked by applications.



\section{Conclusions and future work}
\label{sec:discussion}

In this paper, we presented an approach for automated intrusion detection based on statistical learning methods. We evaluated it for a realistic use case using measurements from an in-house testbed. The results, which are summarized in Section \ref{subsec:evaluation}, show that a defender can predict an attack with high accuracy if sufficient training data is available. Depending on the specific question we are interested in, the nature and amount of the training data, and the availability of computing resources, the findings suggest we use either HMM, LSTM, or RFC. Our results further suggest that our approach can be used to extend the capabilities of traditional intrusion detection systems like SNORT.
Regarding future work, we plan to investigate how the prediction accuracy depends on the type and intensity of background activities by regular users or by maintenance processes. Further, recognizing that an attacker can follow a variety of possible attack sequences to achieve a specific goal, we plan to study prediction methods that are applicable to a large family of attack sequences as a possible extension of this work.


\section{Acknowledgements}
\label{sec:ack}
The authors are grateful to Kim Hammar and Forough Shahab for fruitful discussions around this work and for their comments on an earlier version of this paper. This research has been supported by Digital Futures and the WASP NEST program through project $\mathrm{AIR}^{2}$. 


\bibliographystyle{IEEEtran}
\bibliography{noms2023}

@inproceedings{hammar2022learning,
  author={{K.  Hammar and R.  Stadler}},
  title = {Learning Security Strategies through Game Play and Optimal Stopping},
  booktitle = {Proceedings of the ML4Cyber workshop,
               {ICML} 2022, Baltimore, USA, July
               17-23, 2022},
  publisher = {{PMLR}},
  year      = {2022}
}

@article{rabiner1989tutorial,
  title={A tutorial on hidden Markov models and selected applications in speech recognition},
  author={Rabiner, Lawrence R},
  journal={Proceedings of the IEEE},
  volume={77},
  number={2},
  pages={257--286},
  year={1989},
  publisher={Ieee}
}

@inproceedings{snort,
author = {Roesch, Martin},
title = {Snort - Lightweight Intrusion Detection for Networks},
year = 1999,
publisher = {USENIX Association},
address = {USA},
booktitle = {Proceedings of the 13th USENIX Conference on System Administration},
pages = {229–238},
numpages = 10,
location = {Seattle, Washington},
series = {LISA '99}
}

@article{ids_survey,
	Author = {Khraisat and others},
	Da = {2019/07/17},
	Doi = {10.1186/s42400-019-0038-7},
	Id = {Khraisat2019},
	Isbn = {2523-3246},
	Journal = {Cybersecurity},
	Number = {1},
	Pages = {20},
	Title = {Survey of intrusion detection systems: techniques, datasets and challenges},
	Ty = {JOUR},
	Url = {https://doi.org/10.1186/s42400-019-0038-7},
	Volume = {2},
	Year = {2019}}

@article{srivastava2008credit,
  title={Credit card fraud detection using hidden Markov model},
  author={Srivastava, Abhinav and Kundu, Amlan and Sural, Shamik and Majumdar, Arun},
  journal={IEEE Transactions on dependable and secure computing},
  volume={5},
  number={1},
  pages={37--48},
  year={2008},
  publisher={IEEE}
}

@article{annachhatre2015hidden,
  title={Hidden Markov models for malware classification},
  author={Annachhatre, Chinmayee and Austin, Thomas H and Stamp, Mark},
  journal={Journal of Computer Virology and Hacking Techniques},
  volume={11},
  number={2},
  pages={59--73},
  year={2015},
  publisher={Springer}
}

@article{holgado2017real,
  title={Real-time multistep attack prediction based on hidden markov models},
  author={Holgado, Pilar and Villagr{\'a}, V{\'\i}ctor A and Vazquez, Luis},
  journal={IEEE Transactions on Dependable and Secure Computing},
  volume={17},
  number={1},
  pages={134--147},
  year={2017},
  publisher={IEEE}
}

@article{ghafir2019hidden,
  title={Hidden Markov models and alert correlations for the prediction of advanced persistent threats},
  author={Ghafir, Ibrahim and Kyriakopoulos, Konstantinos G and Lambotharan, Sangarapillai and Aparicio-Navarro, Francisco J and AsSadhan, Basil and Binsalleeh, Hamad and Diab, Diab M},
  journal={IEEE Access},
  volume={7},
  pages={99508--99520},
  year={2019},
  publisher={IEEE}
}

@inproceedings{zhao2017software,
  title={Software abnormal behavior detection based on Hidden Markov Model},
  author={Zhao, Jingling and Huang, Guoxiao and Liu, Tianyu and Cui, Baojiang},
  booktitle={International Conference on Innovative Mobile and Internet Services in Ubiquitous Computing},
  pages={929--940},
  year={2017},
  organization={Springer}
}

@article{shawly2019architectures,
  title={Architectures for detecting interleaved multi-stage network attacks using hidden Markov models},
  author={Shawly, Tawfeeq and Elghariani, Ali and Kobes, Jason and Ghafoor, Arif},
  journal={IEEE Transactions on Dependable and Secure Computing},
  volume={18},
  number={5},
  pages={2316--2330},
  year={2019},
  publisher={IEEE}
}

@article{aoudni2022cloud,
  title={Cloud security based attack detection using transductive learning integrated with Hidden Markov Model},
  author={Aoudni, Yassine and Donald, Cecil and Farouk, Ahmed and Sahay, Kishan Bhushan and Babu, D Vijendra and Tripathi, Vikas and Dhabliya, Dharmesh},
  journal={Pattern Recognition Letters},
  volume={157},
  pages={16--26},
  year={2022},
  publisher={Elsevier}
}

@article{dass2021,
  title={Attack Prediction using Hidden Markov Model},
  author={Dass, Shuvalaxmi and Datta, Prerit and Namin, Akbar Siami},
  journal={arXiv preprint arXiv:2106.02012},
  year={2021}
}

@inproceedings{zan2009hidden,
  title={A Hidden Markov Model based framework for tracking and predicting of attack intention},
  author={Zan, Xin and Gao, Feng and Han, Jiuqiang and Sun, Yu},
  booktitle={2009 International Conference on Multimedia Information Networking and Security},
  volume={2},
  pages={498--501},
  year={2009},
  organization={IEEE}
}

@article{deshmukh2019attacker,
  title={Attacker behaviour profiling using stochastic ensemble of hidden Markov models},
  author={Deshmukh, Soham and Rade, Rahul and Kazi, Dr and others},
  journal={arXiv preprint arXiv:1905.11824},
  year={2019}
}

@article{arnes2006using,
  title={Using hidden markov models to evaluate the risks of intrusions: System architecture and model validation},
  author={ARNES, Andr{\'e} and VALEUR, Fredrik and VIGNA, Giovanni and KEMMERER, Richard A},
  journal={Lecture notes in computer science},
  pages={145--164},
  year={2006},
  publisher={Springer}
}

@article{zhang2014application,
  title={The application of baum-welch algorithm in multistep attack},
  author={Zhang, Yanxue and Zhao, Dongmei and Liu, Jinxing},
  journal={The Scientific World Journal},
  volume={2014},
  year={2014},
  publisher={Hindawi}
}

@article{forney1973viterbi,
  title={The viterbi algorithm},
  author={Forney, G David},
  journal={Proceedings of the IEEE},
  volume={61},
  number={3},
  pages={268--278},
  year={1973},
  publisher={Ieee}
}

@article{rabiner1986introduction,
  title={An introduction to hidden Markov models},
  author={Rabiner, Lawrence and Juang, Biinghwang},
  journal={ieee assp magazine},
  volume={3},
  number={1},
  pages={4--16},
  year={1986},
  publisher={IEEE}
}

@article{sharafaldin2018toward,
  title={Toward generating a new intrusion detection dataset and intrusion traffic characterization.},
  author={Sharafaldin, Iman and Lashkari, Arash Habibi and Ghorbani, Ali A},
  journal={ICISSp},
  volume={1},
  pages={108--116},
  year={2018}
}

@inproceedings{wang2004modeling,
  title={Modeling program behaviors by hidden Markov models for intrusion detection},
  author={Wang, Wei and Guan, Xiao-Hong and Zhang, Xiang-Liang},
  booktitle={Proceedings of 2004 International Conference on Machine Learning and Cybernetics (IEEE Cat. No. 04EX826)},
  volume={5},
  pages={2830--2835},
  year={2004},
  organization={IEEE}
}

@inproceedings{jiang2005novel,
  title={A novel intrusions detection method based on HMM embedded neural network},
  author={Jiang, Weijin and Xu, Yusheng and Xu, Yuhui},
  booktitle={Advances in Natural Computation: First International Conference, ICNC 2005, Changsha, China, August 27-29, 2005, Proceedings, Part I 1},
  pages={139--148},
  year={2005},
  organization={Springer}
}

@article{chen2016anomaly,
  title={Anomaly network intrusion detection using hidden Markov model},
  author={Chen, Chia-Mei and Guan, Dah-Jyh and Huang, Yu-Zhi and Ou, Ya-Hui},
  journal={Int. J. Innov. Comput. Inform. Control},
  volume={12},
  pages={569--580},
  year={2016}
}

@article{hammar2023learning,
  title={Learning near-optimal intrusion responses against dynamic attackers},
  author={Hammar, Kim and Stadler, Rolf},
  journal={IEEE Transactions on Network and Service Management},
  year={2023}
}

@misc{cve,
 title = {Mitre inc},
 year = {2023},
 url = {http://cve.mitre.org/},
 institution = {},
}

@misc{darpa,
 title = {Darpa - intrusion detection evaluation dataset},
 year = {2023},
 url = {https://www.ll.mit.edu/r-d/datasets/2000-darpa-intrusion-detection-scenario-specific-datasets},
 institution = {},
}

@misc{dvwa,
 title = {Damn vulnerable web application (dvwa)},
 year = {2023},
 url = { https://github.com/digininja/DVWA},
 institution = {},
}

@inproceedings{sperotto2009hidden,
  title={Hidden Markov Model modeling of SSH brute-force attacks},
  author={Sperotto, Anna and Sadre, Ramin and de Boer, Pieter-Tjerk and Pras, Aiko},
  booktitle={Integrated Management of Systems, Services, Processes and People in IT: 20th IFIP/IEEE International Workshop on Distributed Systems: Operations and Management, DSOM 2009, Venice, Italy, October 27-28, 2009. Proceedings 20},
  pages={164--176},
  year={2009},
  organization={Springer}
}

@article{hochreiter1997long,
  title={Long short-term memory},
  author={Hochreiter, Sepp and Schmidhuber, J{\"u}rgen},
  journal={Neural computation},
  volume={9},
  number={8},
  pages={1735--1780},
  year={1997},
  publisher={MIT press}
}

@article{breiman2001random,
  title={Random forests},
  author={Breiman, Leo},
  journal={Machine learning},
  volume={45},
  pages={5--32},
  year={2001},
  publisher={Springer}
}

@misc{TBcomplexity,
 author = {scikit-learn developers},
 year = {2007-2019},
 title = {Decision Trees},
 url = {http://mldata.org/repository/data/viewslug/realm-cnsm2015-vod-traces/},
 institution = {},
}

@misc{RFclassifier,
 author = {{Sklearn Developers}},
 title = {RandomForestClassifier},
 howpublished = {[Online]. Available at: \url{https://scikit-learn.org/stable/modules/generated/sklearn.ensemble.RandomForestClassifier.html}, Accessed on: October 5, 2023.},
 institution = {},
}

@misc{LSTM,
 author = {{Keras Developers}},
 title = {Code examples},
 howpublished = {[Online]. Available at: \url{https://www.tensorflow.org/guide/
keras/rnn}, Accessed on: October 5, 2023.},
 institution = {},
}

@inproceedings{rotman2021shuffling,
  title={Shuffling recurrent neural networks},
  author={Rotman, Michael and Wolf, Lior},
  booktitle={Proceedings of the AAAI Conference on Artificial Intelligence},
  volume={35},
  number={11},
  pages={9428--9435},
  year={2021}
}

@article{staudemeyer2015applying,
  title={Applying long short-term memory recurrent neural networks to intrusion detection},
  author={Staudemeyer, Ralf C},
  journal={South African Computer Journal},
  volume={56},
  number={1},
  pages={136--154},
  year={2015},
  publisher={South African Computer Society (SAICSIT)}
}

@article{diro2018leveraging,
  title={Leveraging LSTM networks for attack detection in fog-to-things communications},
  author={Diro, Abebe and Chilamkurti, Naveen},
  journal={IEEE Communications Magazine},
  volume={56},
  number={9},
  pages={124--130},
  year={2018},
  publisher={IEEE}
}

@inproceedings{althubiti2018lstm,
  title={LSTM for anomaly-based network intrusion detection},
  author={Althubiti, Sara A and Jones, Eric Marcell and Roy, Kaushik},
  booktitle={2018 28th International telecommunication networks and applications conference (ITNAC)},
  pages={1--3},
  year={2018},
  organization={IEEE}
}

@article{laghrissi2021intrusion,
  title={Intrusion detection systems using long short-term memory (LSTM)},
  author={Laghrissi, FatimaEzzahra and Douzi, Samira and Douzi, Khadija and Hssina, Badr},
  journal={Journal of Big Data},
  volume={8},
  number={1},
  pages={65},
  year={2021},
  publisher={Springer}
}

@article{zhou2021detecting,
  title={Detecting multi-stage attacks using sequence-to-sequence model},
  author={Zhou, Peng and Zhou, Gongyan and Wu, Dakui and Fei, Minrui},
  journal={Computers \& Security},
  volume={105},
  pages={102203},
  year={2021},
  publisher={Elsevier}
}

@inproceedings{stefanova2017network,
  title={Network attribute selection, classification and accuracy (NASCA) procedure for intrusion detection systems},
  author={Stefanova, Zheni and Ramachandran, Kandethody},
  booktitle={2017 IEEE International Symposium on Technologies for Homeland Security (HST)},
  pages={1--7},
  year={2017},
  organization={IEEE}
}

@inproceedings{lashkari2017characterization,
  title={Characterization of tor traffic using time based features},
  author={Lashkari, Arash Habibi and Gil, Gerard Draper and Mamun, Mohammad Saiful Islam and Ghorbani, Ali A},
  booktitle={International Conference on Information Systems Security and Privacy},
  volume={2},
  pages={253--262},
  year={2017},
  organization={SciTePress}
}

@inproceedings{mcelwee2017active,
  title={Active learning intrusion detection using k-means clustering selection},
  author={McElwee, Steven},
  booktitle={SoutheastCon 2017},
  pages={1--7},
  year={2017},
  organization={IEEE}
}

@inproceedings{choubisa2022simple,
  title={A simple and robust approach of random forest for intrusion detection system in cyber security},
  author={Choubisa, Manish and Doshi, Ruchi and Khatri, Narendra and Hiran, Kamal Kant},
  booktitle={2022 International Conference on IoT and Blockchain Technology (ICIBT)},
  pages={1--5},
  year={2022},
  organization={IEEE}
}

@article{gupta2016framework,
  title={A framework for fast and efficient cyber security network intrusion detection using apache spark},
  author={Gupta, Govind P and Kulariya, Manish},
  journal={Procedia Computer Science},
  volume={93},
  pages={824--831},
  year={2016},
  publisher={Elsevier}
}

@article{belavagi2016performance,
  title={Performance evaluation of supervised machine learning algorithms for intrusion detection},
  author={Belavagi, Manjula C and Muniyal, Balachandra},
  journal={Procedia Computer Science},
  volume={89},
  pages={117--123},
  year={2016},
  publisher={Elsevier}
}

@inproceedings{da2017detecting,
  title={Detecting mobile botnets through machine learning and system calls analysis},
  author={da Costa, Victor GT and Barbon, Sylvio and Miani, Rodrigo S and Rodrigues, Joel JPC and Zarpel{\~a}o, Bruno B},
  booktitle={2017 IEEE International Conference on Communications (ICC)},
  pages={1--6},
  year={2017},
  organization={IEEE}
}

@inproceedings{sai2019advance,
  title={Advance persistent threat detection using long short term memory (LSTM) neural networks},
  author={Sai Charan, PV and Gireesh Kumar, T and Mohan Anand, P},
  booktitle={Emerging Technologies in Computer Engineering: Microservices in Big Data Analytics: Second International Conference, ICETCE 2019, Jaipur, India, February 1--2, 2019, Revised Selected Papers 2},
  pages={45--54},
  year={2019},
  organization={Springer}
}

\end{document}